\documentclass[journal]{IEEEtran}
\usepackage[T1]{fontenc}
\usepackage{color}
\usepackage{flushend}
\usepackage{lipsum}
\usepackage{multirow}
\usepackage{amsmath}
\usepackage{booktabs}
\usepackage{tabu}
\usepackage[cmintegrals]{newtxmath}
\usepackage{bm}
\usepackage{indentfirst}
         
\usepackage{amsmath,amssymb}
\usepackage{array}
\usepackage{mathrsfs}
\usepackage{cite}
\usepackage{tikz}
\usepackage{color}

\usepackage{url}
\usepackage{hyperref}
\usepackage{todonotes}
\usepackage{color}

\begin{document}

\title{Detect and Locate: Exposing Face Manipulation by Semantic- and Noise-level Telltales}

\author{Chenqi~Kong,
        Baoliang~Chen,
        Haoliang~Li,~\IEEEmembership{Member,~IEEE}, 
        Shiqi~Wang,~\IEEEmembership{Senior Member,~IEEE},\\
        Anderson~Rocha,~\IEEEmembership{Senior Member,~IEEE},
        and~Sam~Kwong,~\IEEEmembership{Fellow,~IEEE}%
\thanks{C. Kong, B. Chen, S. Wang and S. Kwong are with the Department of Computer Science, City University of Hong Kong, Hong Kong, China. (email: cqkong2-c@my.cityu.edu.hk; blchen6-c@my.cityu.edu.hk; shiqwang@cityu.edu.hk; cssamk@cityu.edu.hk).}
\thanks{H. Li is with the Department of Electrical Engineering, City University of Hong Kong, Hong Kong, China. (email: haoliang.li@cityu.edu.hk).}
\thanks{A. Rocha is with the Artificial Intelligence Lab. (\texttt{Recod.ai}) at the University of Campinas, Campinas 13084-851, Brazil (e-mail: arrocha@unicamp.br), URL: \url{http://recod.ai}}
\thanks{S. Wang is the corresponding author.}
}

\markboth{}%
{Shell \MakeLowercase{\textit{et al.}}: Bare Demo of IEEEtran.cls for IEEE Communications Society Journals}

\maketitle

\begin{abstract}
The technological advancements of deep learning have enabled sophisticated face manipulation schemes, raising severe trust issues and security concerns in modern society. Generally speaking, detecting manipulated faces and locating the potentially altered regions are challenging tasks. Herein, we propose a conceptually simple but effective method to efficiently detect forged faces in an image while simultaneously locating the manipulated regions. The proposed scheme relies on a segmentation map that delivers meaningful high-level semantic information clues about the image. Furthermore, a noise map is estimated, playing a complementary role in capturing low-level clues and subsequently empowering decision-making. Finally, the features from these two modules are combined to distinguish fake faces. Extensive experiments show that the proposed model achieves state-of-the-art detection accuracy and remarkable localization performance.  
\end{abstract}


\IEEEpeerreviewmaketitle
\section{Introduction}
\IEEEPARstart{S}{uccessful} computer vision technologies have recently also fueled face manipulation strategies with a series of methods proposed, including Deepfakes \cite{hm16_20}, Face2Face \cite{dfcode}, FaceSwap\cite{thies2016face2face} and NeuralTextures \cite{thies2019deferred}. The fake content created by these methods is becoming increasingly realistic. Falsified video content, involving both manipulated identity and edited expression, raises various disconcerting problems within widespread social media, such as identity theft, fake news dissemination, and fraud. Even worse, most existing face manipulation technologies are disclosed online and easy to implement. Even a non-expert person without any prior professional skills can create high-quality fake content. It cannot be denied that these manipulation techniques and off-the-shelf commercial products (e.g., FakeApp \cite{fakeapp}) bring certain positive impacts to the film-making industry and other entertainment applications. However, face manipulation techniques are highly likely to be misused, and the potential malicious applications are  hazardous. 
\begin{figure}[htb]
\centering
\includegraphics[scale=0.20]{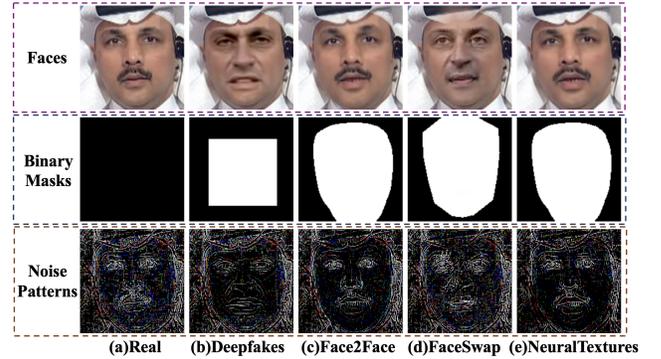}
\caption{Illustration of an authentic and four corresponding manipulated faces, including the face images (top row), binary masks (middle row), and noise patterns (bottom row). In the binary mask, the black pixels denote the masking label of the pristine face, and the white pixels represent the region that has been tampered with.
}
\label{teaser}
\end{figure}
As such, there has been an exponential increase in the demand for efficient face manipulation
detection methods to counteract its dangerous impacts.

Despite the demonstrated success in developing detection methods, the advancement of face manipulation techniques still poses a grand challenge.
The situation worsens when manipulated videos are compressed before distribution. One general detection methodology includes leveraging high-level semantic clues (e.g., lack of eye blinking~\cite{li2018ictu}, head pose inconsistency~\cite{yang2019exposing}, and face splicing~\cite{li2020face}). Such methods may ignore low-level signal variations in manipulated videos, resulting in limitations in terms of robustness and effectiveness. This has motivated the development of frequency-domain features
~\cite{qian2020thinking, liu2021spatial, masi2020two}. Manipulation region localization \cite{dang2020detection, li2020face, du2020towards} is another task of paramount importance in face forensics as it is pivotal to unveil the intention of a forger. However, this aspect has been largely ignored in most existing methods. In this paper, we conduct the manipulation localization at the pixel level. To produce manipulated faces, most face manipulation methods need to blend the altered face region into a background image. As such, the pixels in the manipulated face region will be altered due to the synthesis algorithm and the blending process.
Another concern of face manipulation is the degradation of low-quality videos' detection accuracy due to the limited generalization capability across different quality levels. This could primarily affect the deployment of detection methods in practical applications.

In this vein, we aim to develop an method to expose face manipulation with three desired properties,
\begin{itemize}
    \item \textbf{High detection accuracy}: It must achieve high accuracy in detecting manipulated faces. 
    \item \textbf{Outstanding localization ability}: It must be able to locate the manipulated regions precisely. 
    \item \textbf{Flexibility}: It must be robust to videos in different quality levels. 
\end{itemize}
We introduce a two-stream multi-scale framework based on the fusion of semantic-level and noise-level features, achieving the three desired properties mentioned above, simultaneously.  

As Fig.~\ref{teaser} depicts, the semantic-level guidance (middle row) reveals the location of the manipulated face, rooted at the generally adopted procedure in face manipulation by blending the new face with a background. On the other hand, the noise-level guidance originates from noise patterns (bottom row), and observers may find that the noise pattern of the pristine face is more consistent and homogeneous than others. As such, semantic masks and noise patterns are employed as guiding labels to supervise the model's training. Besides, we extract three feature maps from the shallow, middle and deep layers of the convolutional neural network backbone, such that the extracted multi-scale features can carry information from low, middle, and high levels. Finally, we combine the semantic-level and noise-level features to check authenticity and locate the manipulated region of the input face. Extensive experiments demonstrate that the proposed framework achieves consistent performance improvement compared to state-of-the-art methods, holds promise for both high and low-quality videos, and delivers more information regarding the manipulated regions.

\section{Related Work}

\subsection{Face Manipulation and Detection}
Generally speaking, face manipulation methods can be classified into the following two categories: identity swap and facial expression manipulation. In particular, Deepfakes~\cite{hm16_20} and FaceSwap~\cite{dfcode} are two prominent representatives of face identity swap techniques. Face2Face~\cite{thies2016face2face} and NeuralTextures~\cite{thies2019deferred} are two popular facial expression editing methods. Deepfakes and NeuralTextures are learning-based methods among those four face manipulation techniques, while FaceSwap and Face2Face are computer graphics-based methods. 

Due to the potential malicious applications of face manipulation techniques, numerous face forgery detectors have been proposed. For example, Li \textit{et al.}~\cite{li2020face} proposed the Face X-ray to detect the blending boundary of the fake face, which also reveals the manipulated boundaries. Chintha~\textit{et al.}~\cite{chintha2020recurrent} applied a recurrent structure to capture spatial and temporal information of manipulated videos. In
~\cite{songsri2019complement} and \cite{dang2020detection}, the authors designed two models to perform face forensics detection and localization. Li~\textit{et al.}~\cite{li2020sharp} proposed a sharp-MIL (Multiple Instance Learning) to perform face forgery detection on videos. Qian \textit{et al.}~\cite{qian2020thinking} introduced frequency analysis into face forgery detection and achieved promising detection results. Masi
~\textit{et al.} \cite{masi2020two} designed a two-branch recurrent network to detect face manipulation. None of the above methods considered combining semantic and noise signatures to detect forgeries and basically all of them suffer from performance drop when dealing with low-quality videos or cross-dataset evaluations.

\subsection{Binary Mask Supervision}
Image manipulation segmentation focuses on regions which are potentially manipulated~\cite{bappy2017exploiting} and often relies upon guiding binary masks.
The ground-truth masks are generated by marking forged regions as `1' and the pristine (un-forged) regions as `0'. In image forensics, Zhou~\textit{et al.} \cite{zhou2018learning} fused the features from the RGB stream and the noise stream to capture the manipulated boundaries as well as the noise inconsistencies between the authentic and manipulated regions, and the features are then used to locate the potentially manipulated regions. Bappy \textit{et al.} \cite{bappy2017exploiting} designed a hybrid CNN-LSTM model to capture boundary discrepancy features and perform image manipulation localization. Zhou \textit{et al.} \cite{zhou2020generate} employed a novel generator to perform data augmentation and trained a model to complete image forgery localization. The binary mask guidance has also been widely used in face anti-spoofing. In \cite{george2019deep, liu2019deep, sun2020face}, binary masks are employed to perform pixel-wise supervision during the training process. These models aim to discover arbitrary cues that can identify the live (pristine) and spoofed (manipulated) faces. In face manipulation detection, Du
~\textit{et al.} \cite{du2020towards} designed a locality-aware autoencoder to enforce the model to learn the feature representation of the forgery region, which boosts the generalization capability. Dang \textit{et al.} \cite{dang2020detection} utilized the binary mask and an attention mechanism to make the extracted feature maps focus on manipulated regions, which further improves the detection performance. Li \textit{et al.} \cite{li2020face} designed a more general model to capture the boundary splicing artifacts which can explicitly locate the potentially manipulated facial region. 

Differently from the previous methods, in this paper, we design a multi-scale semantic map prediction module to capture the semantic-level information of input faces and simultaneously perform the localization of face manipulation. The binary masks play a supervision role in the model training process, encouraging the network to learn features that account for such fine-grained manipulations.

\begin{figure*}[h]
\centering
\includegraphics[scale=0.47]{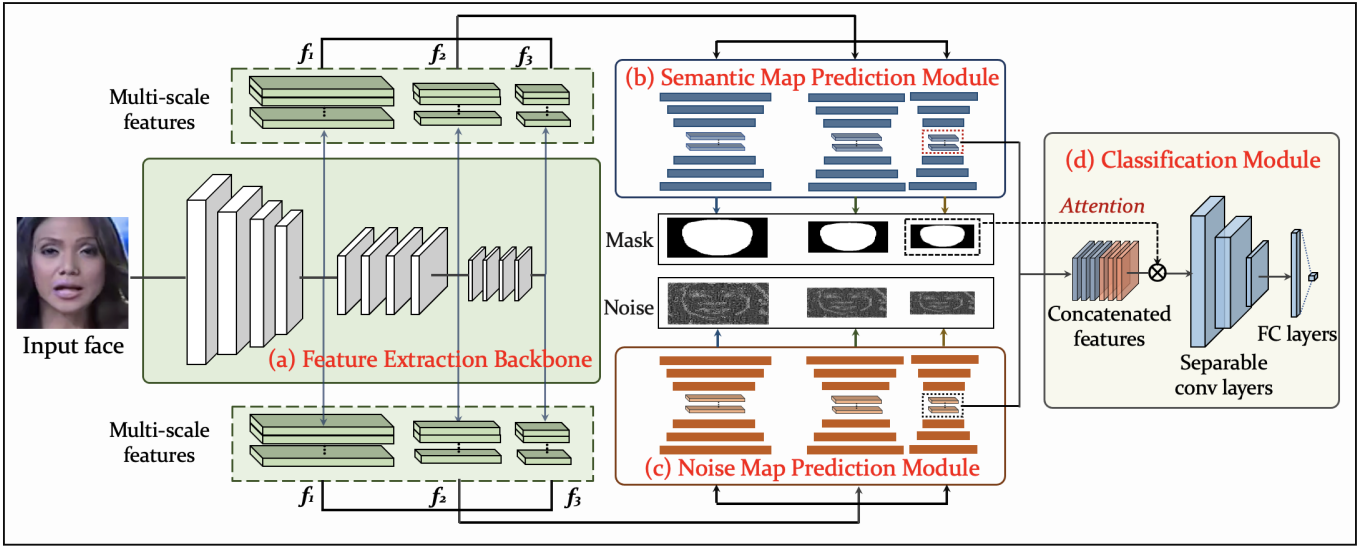}
\caption{Overview of the proposed framework. The face manipulation detection model consists of four components: (a) A feature extraction backbone; (b) A semantic map prediction module; (c) A noise map prediction module; and (d) a classification module. The faces are first forwarded to the feature extraction network backbone, and three feature maps extracted from the shallow, middle, and deep layers are synchronously processed by the semantic map prediction module and the noise map prediction module (the noise map intensity has been enlarged by five times for better visualization). A weighted summation over the three predicted semantic segmentation maps are conducted to perform the face manipulation localization.
The estimated semantic and noise features are combined, and the spatial attention is conducted by element-wisely multiplying the combined features with the last predicted semantic segmentation maps. Finally, the attention-based feature maps are forwarded to the classification module for final decision-making.}
\label{overview}
\end{figure*}

\subsection{Noise Modeling}
Pioneering work on noise modeling for image forensics date back to 2004, where inconsistencies at the noise level have been leveraged to expose splicing artifacts of manipulated images \cite{popescu2004statistical, mahdian2009using, lyu2014exposing}. More recently, photo-response non-uniformity (PRNU) \cite{fridrich2006digital} noise patterns have been widely studied in multimedia forensics. In general, PRNU can also be regarded as the device fingerprint. Korus \textit{et al.} \cite{korus2016multi} proposed a multi-scale PRNU-based scheme to tackle the tampering localization problem. Quan \textit{et al.} \cite{quan2020addressing} focused on the correlation between the noise residual with the PRNU pattern for forgery detection. Cozzolino \textit{et al.} \cite{cozzolino2019noiseprint} applied a noise extraction Siamese network \cite{zagoruyko2015learning} to achieve manipulation detection and forgery localization. 

Noise patterns are also regarded as key clues in face anti-spoofing. More specifically, Jourabloo \textit{et al.} \cite{jourabloo2018face} designed a CNN to decompose a spoofing face into the spoofing noise and a live face, and the spoofing noise is further used to perform the live/spoofing identification. Ren \textit{et al.} \cite{ren2020face} designed a noise-attention network architecture to extract noise-features for live/spoofing face classification. 

While noise signature has been widely used in image forensics and face anti-spoofing tasks, how the noise signature benefits face manipulation detection, especially the fake face contents generated by deep learning technologies, has not been investigated yet. In this work, we argue that some key cues remain in the noise maps of manipulated faces. Therefore, a noise map prediction module is designed to extract the noise pattern of corresponding input faces, and the noise feature is subsequently fed to a classification module for face manipulation detection.

\section{Proposed Method}
This section presents a method to simultaneously detect and locate face manipulated regions. It relies upon high-level semantic information clues about an image  combined with noise low-level features.  The method consists of four components: (a) a feature extraction backbone; (b) a semantic map prediction module; (c) a noise map prediction module; and (d) a classification module. 

\subsection{Overall Framework}
We propose a two-stream multi-scale framework for face forgery detection and localization, and the architecture is illustrated in Fig.~\ref{overview}. In particular,  we denote the training dataset as $D=\{F_{i}, M_{i}, n_{i}, c_{i}\}_{i=1}^{N}$, where $N$ is the total number of training samples.  The dataset consists of the following four components: the $i$th input face {$F_i$}, its corresponding binary mask label {$M_i$}, noise pattern label {$n_i$} and  ground-truth binary label of forgery {$c_i$}. We employ the Xception network \cite{chollet2017xception} as the backbone to learn multi-scale feature maps based on two insights. Other networks could also be used without loss of generality. 

First,  multi-scale features enable the model to learn both semantic and geometric information as features from different layers contribute to different receptive fields. Second, multi-scale supervision leads the model to focus on the valuable information from the beginning (i.e., learning face-forgery-specific information at the shallow layer), which benefits face manipulation detection and localization performance. 

The semantic map prediction module and the noise map prediction module synchronously processes the multi-scale feature maps, where the outputs are guided by binary mask label $M_i$ and noise pattern label $n_i$, respectively. The semantic map prediction module is trained to capture the high-level semantic information from the extracted multi-scale features where the estimated semantic map indicates the potentially manipulated regions in the corresponding input faces. 

The three predicted semantic segmentation maps are leveraged to perform forgery localization. The noise map prediction module is further proposed to enforce the multi-scale features to focus on image content and pay attention to content-irrelevant low-level information. The estimated noise maps contain rich high-frequency cues that might expose noise artifacts in manipulated faces by noise modeling. 

We combine the last semantic features and the last noise features, and the spatial attention is sequentially conducted by element-wisely multiplying the concatenated features with the last predicted semantic segmentation maps. Finally, the attention-based features are forwarded to the classification module to identify the authenticity of input faces. 

The overall objective function consists of three components: 
\begin{equation}
     L = L_{c} + {\lambda}_{1} L_{n} + {\lambda}_{2} L_{b},
\end{equation}
where {$L_{c}$}, {$L_{n}$}, and {$L_{b}$} denote final classification loss, noise map prediction loss, and semantic map estimation loss, respectively. {${\lambda}_{1}$} and {${\lambda}_{2}$} are hyper-parameters to weigh the loss components. {$L_{c}$} is the cross-entropy loss between the prediction result {$\hat{c}_{i}$} and ground truth label {$c_{i}$}:
\begin{equation}
     L_{c}= -\frac{1}{N}\sum_{{i=1}}^{N}
     (c_{i}\log\hat{c}_{i}+(1-c_{i})\log(1-\hat{c}_{i})),
\end{equation}
We better describe {$L_{n}$} and {$L_{b}$} in the following subsections. 

\subsection{Semantic Map Prediction Module}
Binary mask supervision has been widely used to tackle various forensics problems. In this paper, we generate the ground-truth mask for each face in the training stage and use the mask to supervise the training of the proposed model. The benefits of the semantic map prediction module are mainly threefold: 1) it enables the model to localize manipulated regions, providing evidence to show whether the input faces have been manipulated; 2) it constrains the model to focus on  manipulated regions, leading the model to achieve a better manipulation detection performance; and 3) different manipulation approaches tend to have different binary mask shapes, thus providing auxiliary information to complete multi-class classification (more details are elaborated in Sec.IV-C.4)). 

 The semantic map prediction module takes the extracted multi-scale features as inputs, where the outputs are the estimated semantic segmentation maps. In particular, we apply the depth-wise separable convolution block as our semantic map prediction module as it has been shown to be effective in different types of problems \cite{chollet2017xception}. For the output of the semantic map prediction module, each pixel value indicates the probability that the corresponding receptive field is a manipulated region in the input face. Herein, we conduct a weighted summation strategy over the estimated semantic maps in multi-scale to determine the final manipulation localization map of the input faces. 
 
We use videos in FaceForensics++ \cite{rossler2019faceforensics++} as our training set as it provides a manipulation mask for each manipulated face, supervising the training of the semantic map prediction module. Training the semantic map prediction module can tackle a binary classification problem, including the manipulated region and the non-manipulated area. Furthermore, the cross-entropy loss is leveraged to supervise the estimation of semantic segmentation maps:

\begin{small}
\begin{equation}
     L_{b}= -\frac{1}{N}\sum_{{i=1}}^{N}\sum_{{j}}\frac{1}{N_{p}}\sum_{{x,y}} 
     (M^{x,y}_{i,j}\log\hat{M}^{x,y}_{i,j}+(1-M^{x,y}_{i,j})\log(1-\hat{M}^{x,y}_{i,j})),
\end{equation}
\end{small}

\begin{figure}[h]
\centering
\includegraphics[scale=0.45]{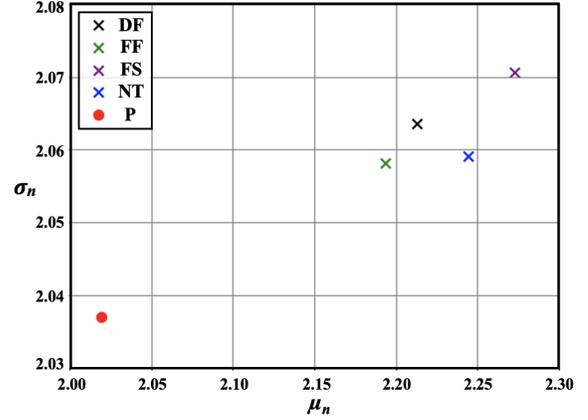}
\caption{The statistical analysis of faces from four manipulation methods and pristine faces. Red color dots represent pristine (P) face images, and the remaining four colors indicate the four manipulation methods: Deepfakes(DF), Face2Face(FF), FaceSwap(FS), and NeuralTextures(NT).}
\label{noise}
\end{figure}

\noindent where {$\hat{M}^{x,y}_{i,j}$} and {$M^{x,y}_{i,j}$} represent the estimated and ground truth semantic maps, and the sizes of all ground truth maps have been aligned to the corresponding predicted maps. The parameters {$x$} and {$y$} denote the pixel location; {$i$} specifies {$i$}th input face; {$j=\{shallow, middle, deep\}$} denotes the mask predicted by the specific feature layer; {$N_{p}$} is the total pixel number.

\subsection{Noise Map Prediction Module}
One notable example of forensics-related noise pattern analysis is PRNU, which is the dominant part of the noise in natural images caused by the inhomogeneity of silicon wafers and imperfections during the sensor manufacturing process~\cite{lukas2006digital} and has been widely applied in different multimedia forensics tasks~\cite{lukas2006digital, korus2016multi, cozzolino2019noiseprint} to date. Inspired by the PRNU paradigm, we leverage the wavelet-based filter \cite{mihcak1999spatially} to extract a noise map to explore further the face image forensics artifacts as the low-level clue guidance for face manipulation detection.  The noise map prediction module is designed to make the extracted multi-scale features irrelevant with high-level
semantic content (i.e., facial appearance) and provide a complementary clue for face manipulation detection. 

The design of the noise map prediction module relies upon the assumption that noise patterns of manipulated faces expose two artifacts. First, the noise pattern of the manipulated region tends to expose more abnormal artifacts due to the image processing of the fake face creation process. Second, the statistical discrepancy between the noise distributions of the manipulated face region and the background region introduces noise distribution inconsistencies \cite{cozzolino2019noiseprint, zhou2018learning}. We statistically analyze the noise pattern distributions of four manipulation methods and pristine data in Fig.~\ref{noise}. It can be observed that the mean {$\mu_{n}$} and variance {$\sigma_{n}$} values of pristine faces are much smaller than the values of the other four manipulation methods. Therefore, it is reasonable to employ noise guidance to perform face manipulation detection.

Following previous multimedia forensics methods \cite{lukas2006digital, korus2016multi, cozzolino2019noiseprint}, we apply a wavelet-based filter~\cite{mihcak1999spatially} to extract noise patterns for each face, which can be formulated as:
\begin{figure}[h]
\centering
\includegraphics[scale=0.29]{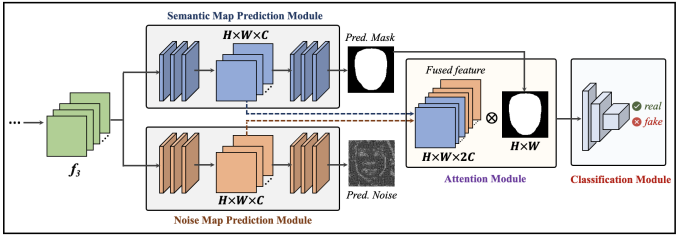}
\caption{Illustration of feature fusion strategy and attention mechanism.}
\label{ATT}
\end{figure}

\begin{figure}[h]
\centering
\includegraphics[scale=0.28]{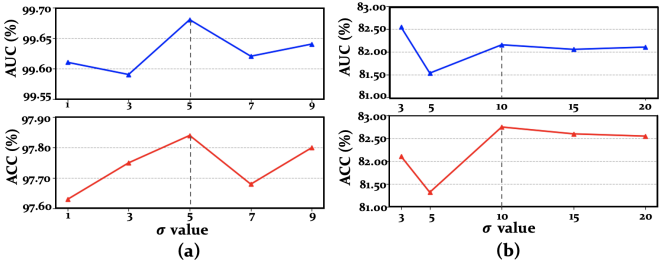}
\caption{The manipulation detection performances of noise prediction module on (a) high-quality faces and (b) low-quality faces. The ground-truth noise maps are generated with different {$\sigma$} values.}
\label{sigma}
\end{figure}

\begin{equation}
     n_{i} = F_{i} - f(F_{i})
\end{equation}
where {$F_{i}$} and {$n_{i}$} represent the $i$th input raw face and the corresponding extracted noise pattern, and {$f(\cdot)$} is defined as the noise filter:
\begin{equation}
     \hat F_{i} =  f(F_{i}) = \frac{\hat\sigma^2_{F_{i}}}{\hat\sigma^2_{F_{i}}+\sigma^2}F_{i}
\end{equation}
where {$\hat F_{i}$} is the denoised face image, {$\sigma^2$} indicates the variance estimation of the additive white Gaussian noise (AWGN), and {$\hat\sigma^2_{F_{i}}$} is the variance of {$\hat F_{i}$}. The estimation details of {$\hat\sigma^2_{F_{i}}$} can be found in \cite{mihcak1999spatially}. 

We can extract the noise residual for each input face as the noise label. We further design a multi-scale noise map prediction module to estimate low-level noise patterns and extract noise-specific features. The extracted noise-specific features aim to expose noise artifacts, serving as auxiliary information to improve the final authentic identification performance. The noise map prediction loss {$L_{n}$} calculates the {$l_{1}$} norm of the pixel-wise difference between predicted noise maps  {$\hat{n}$} and corresponding noise labels {$n$}:
\begin{equation}
     L_{n}= \frac{1}{N} \sum_{{i=1}}^{N}\sum_{{j}}
     \|{n_{i,j}}-{\hat{n}_{i,j}}\|_{1},
\end{equation}
where {$i$} specifies {$i$}th input face; {$j$} denotes the specific feature layer; {$n_{i,j}$} and {$\hat{n}_{i,j}$} represent the ground truth and estimated noise patterns.
This design enables the network to learn the content-irrelevant low-level information and expose noise artifacts. 

\subsection{Feature Fusion and Classification Module}
In this subsection, we present the feature fusion process, attention mechanism, and the final classification module.  We illustrate the feature fusion strategy and the attention mechanism in Fig.~\ref{ATT}. {$f_{3}$} represents the feature extracted from the deep layer of the backbone. First, we feed {$f_{3}$} forward to Semantic Map Prediction Module and Noise Map Prediction Module to obtain the predicted mask and noise map. The high-level semantic feature and low-level noise feature have the same size of H$\times$W$\times$C, and the predicted map size is H$\times$W. Next, we combine the semantic and noise features by concatenating semantic features and low-level features along the channel dimension. Subsequently, we conduct the spatial attention by element-wisely multiplying the fused feature with the predicted mask. Intuitively, the manipulated face region contains abundant abnormalities and artifacts. We aim to mine the artifacts at both semantic- and noise-levels. Therefore, we design the attention mechanism to highlight the manipulated locations and further empower the final binary classification performance. On the other hand, each pixel in the attention map indicates the probability that the corresponding receptive field is a manipulated region in the input face. Element-wisely multiplying the combined features with the attention map can guide the model to focus on the manipulated spatial region and improve binary classification accuracy. Finally, the spatially attended feature is fed to the Classification Module to identify the authenticity of the input query.

\section{Experiments}

\subsection{Data Preparation}
\subsubsection{Dataset} 
In this paper, we conduct the experiments on the challenging FF++ \cite{rossler2019faceforensics++} dataset. FF++ \cite{rossler2019faceforensics++} is a dataset of face manipulation that contains 1,000 pristine videos and 4,000 associated manipulated videos created by four state-of-the-art forgery techniques: Deepfakes \cite{hm16_20}, Face2Face \cite{dfcode}, FaceSwap\cite{thies2016face2face} and NeuralTextures \cite{thies2019deferred}.  Besides, FF++ \cite{rossler2019faceforensics++} provides three quality levels controlled by the quantization parameters (QP) in compression 
for these 5000 videos: raw (QP=0), HQ (high-quality, QP=23), and LQ (low-quality, QP=40). Considering the deployment in real-world application scenarios, we conduct our experiments on both HQ videos and LQ videos. FF++ \cite{rossler2019faceforensics++} also provides the ground truth-masks that indicate the forged regions of manipulated faces, enabling the developments of face forgery localization methods. Following the experimental setting in \cite{rossler2019faceforensics++}, we take 720 videos for training, 140 videos for validation, and 140 videos for testing. We extract 270 frames from each training video and 100 frames from each validation and testing video. Following the  official experimental setting\cite{rossler2019faceforensics++},  during training we augment the number of real faces four times to address the data imbalance between real and fake faces. The summary of the data in training, validation, and testing is listed in Table~\ref{ffdata}. For that we just extract more faces from the training videos. 

\subsubsection{Binary mask generation} FF++ provides a 3D mask for every video frame, and the 3D mask indicates the manipulated region of each face. As such, we generate the ground-truth binary mask for each face according to the boundary location of the 3D mask. The middle row in Fig.~\ref{teaser} illustrates the generated ground-truth masks for real and manipulated faces, where the white pixels denote the manipulated regions, and the black pixels represent the non-manipulated regions (pristine regions). For other data corpora which do not provide mask labels, the alternative way to quickly generate manipulation masks manually can be summarized as two steps: (1) computing the absolute difference between the fake face and the corresponding real face, as most datasets contain the real and fake video pairs, $e.g.,$ CelebDF \cite{li2020celeb}, DFD \cite{dfd}, and DFDC \cite{dolhansky2020deepfake};
(2) setting a threshold value for the absolute difference map and convert it to a binary mask.
This mask generation process can be summarized as follows,

\begin{equation}
     M_{i,j} = \left\{
    \begin{aligned}
    0 & , & |F_{i,j}^{real} - F_{i,j}^{fake}|<threshold, \\
    1 & , & |F_{i,j}^{real} - F_{i,j}^{fake}|\geq threshold.
    \end{aligned}
    \right.
\end{equation}

 \begin{table}
  \caption{Dataset breakdown for training, validation, and testing sets.}
  \label{ffdata}
  \centering
  \renewcommand\arraystretch{1.15}
   \scalebox{0.9}{\begin{tabular}{|c|c|c|c|c|c|c|}
  \hline
  &
  \multicolumn{2}{c|}{\textbf{Training}} &
  \multicolumn{2}{c|}{\textbf{Validation}} &
  \multicolumn{2}{c|}{\textbf{Test}} \\
  \cline{1-7}
    \textbf{Methods} & Videos & Faces & Videos & Faces & Videos & Faces \\
  \hline
  Deepfakes & 720 & 192,748 & 140 & 13,907 & 140 & 14,000 \\
  \hline
  Face2Face & 720 & 194,091 & 140 & 14,000 & 140 & 14,000 \\
  \hline
  FaceSwap & 720 & 193,989 & 140 & 14,000 & 140 & 14,000 \\
  \hline
  NeuralTextures & 720 & 194,004 & 140 & 14,000 & 140 & 14,000 \\\hline
  Real & 720 & 776,912 & 140 & 56,000 & 140 & 56,000 \\
  \hline
\end{tabular}}
\end{table}

\subsubsection{Noise map generation} Following  previous forensics methods~\cite{lukas2006digital, korus2016multi, cozzolino2019noiseprint}, we apply the wavelet-based noise filter~\cite{mihcak1999spatially} to extract the noise map for each input face. The extracted noise maps are regarded as pseudo labels to supervise the training of the noise map prediction module. The hyper-parameter {$\sigma$} as defined in Eq. (5) is the variance estimation of additive white Gaussian noise(AWGN). To determine the best values of {$\sigma$} for high-quality and low-quality faces, we leverage the noise patterns generated with different {$\sigma$} values to supervise the training of the noise map prediction module and the detection performances of high-quality and low-quality faces are illustrated in Fig.~\ref{sigma}. According to the detection results in Fig.~\ref{sigma} (a) and (b), we set the {$\sigma$} values as 5 and 10 for high-quality and low-quality faces. 

\subsection{Implementation Details}
\subsubsection{Training strategy} 
We apply the popular \texttt{dlib} face detector~\cite{king2009dlib} to crop the face regions enlarged by a factor of 1.3. The proposed framework is implemented by Pytorch \cite{paszke2019pytorch}. The model is trained using Adam optimizer \cite{kingma2014adam} with {$\beta_{1}$}=0.9 and {$\beta_{2}$}=0.999. Multi-scale feature layers are extracted after the conv2 , block2, and block11 of the Xception backbone. We release the code here\footnote{\url{https://github.com/ChenqiKONG/Detect_and_Locate}.}.

\begin{figure}[h]
\centering
\includegraphics[scale=0.30]{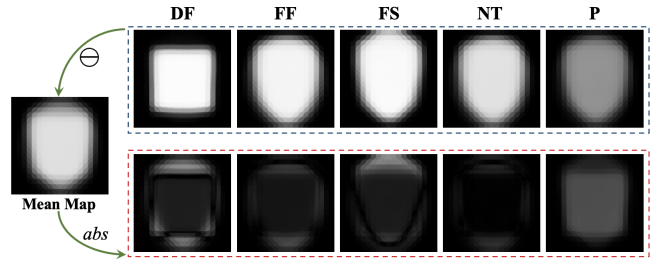}
\caption{Predicted mask maps from the testing set. The mean map on the left represents the average summation over all manipulation localization maps and pristine maps. The top row shows the mean maps of four specific manipulation methods (DF: Deepfakes, FF: Face2Face, FS: FaceSwap, NT: NeuralTextures) and pristine face images (P: Pristine). The maps on the bottom row indicate the absolute differences between the top-row maps and the left mean map.}
\label{Biasmap}
\end{figure}

\begin{figure}[h]
\centering
\includegraphics[scale=0.28]{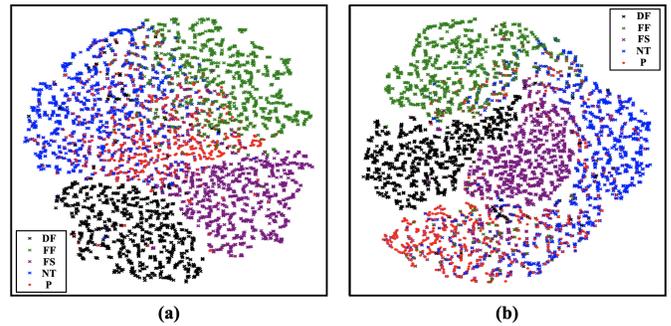}
\caption{T-SNE feature embedding visualization of (a) the Xception baseline and (b) the proposed method on FF++\cite{rossler2019faceforensics++} low-quality (QP=40) dataset in the five-class classification task. Red color dots represent pristine (P) face images, and the rest four colors indicate the four manipulation methods: Deepfakes (DF), Face2Face (FF), FaceSwap (FS), and NeuralTextures (NT). }
\label{tsne}
\end{figure}

We conduct a two-step training strategy: 1) training the Xception backbone and the semantic map prediction module, where the weights of the backbone are initialized with the ImageNet~\cite{deng2009imagenet} weights; and 2) initializing the weights of the Xception backbone and the semantic map prediction module with the weights obtained in Step 1), then training the overall framework in an end-to-end manner. This two-step training strategy can effectively alleviate the over-fitting problem and be more robust to local minima. We set the learning rate and weight decay as 0.0002 and 1e-5, respectively. The model is trained on 2 RTX 2080Ti GPUs with batch size 32.

\subsubsection{Evaluation metrics} 
Following most existing face forgery detection methods, we employ the detection Accuracy rate (ACC) and Area Under the Receiver Operating Characteristic Curve (AUC) as the forgery detection evaluation metrics. ACC is the most straightforward metric in face forgery detection, which directly reflects the detection capability of the detector. AUC is a more objective evaluation metric, which has been widely applied in recent state-of-the-art detection methods, such as Face X-ray \cite{li2020face}, Celeb-DF \cite{li2020celeb}, and Two-branch \cite{masi2020two}. 

This paper takes AUC as our key evaluation metric and reports all detection results at the frame level. We further use the metrics of False Positive Rate (FPR), False Negative Rate (FNR), Equal Error Rate (EER), and Average Precision (AP) to interpret the face manipulation detection results better.

On the other hand, we utilize Intersection over Union (IoU), Pixel-wise Binary Classification Accuracy (PBCA), and Inverse Intersection Non-Containment (IINC) \cite{dang2020detection} as the localization evaluation metrics. IoU is one of the standard metrics in semantic segmentation and has been widely used in numerous semantic segmentation tasks \cite{long2015fully, noh2015learning}. The definition of IoU is illustrated as follows,
\begin{equation}
     IoU = \frac{\hat{M} \cap M}{\hat{M} \cup M}.
\end{equation}
The numerator calculates the area of overlap between the predicted mask {$\hat{M}$} and the ground truth mask {$M$}, and the denominator represents the area encompassed by both the predicted mask and the ground truth mask. In addition to IoU, PBCA and IINC proposed in \cite{dang2020detection}, are two metrics for evaluating face manipulation localization performance. PBCA calculates the mask prediction accuracy 
at the pixel level. IINC measures the non-overlap ratio of both predicted map and ground truth map. Thus it is a more robust and coherent metric in evaluating the localization results. 
As such, IoU, PBCA, and IINC are employed to evaluate the forgery localization accuracy.

\subsection{Manipulation Detection}
We compare the proposed method with previous detection methods in terms of ACC, EER, and AUC.  

\subsubsection{Detection performance on all manipulation methods} Table~\ref{eval_all} reports the detection results of the proposed framework and previous methods trained on the FF++ dataset. We use the bold font to highlight the best results while underlining the second-best results among all listed methods. HQ and LQ indicate the high-quality and low-quality data. Comparing with the most recent work \cite{nirkin2021deepfake}, the proposed model achieves a significant ACC score improvement, going from 80.18\% $\rightarrow$ 84.84\%. Moreover, the overall detection performance of the proposed method outperforms all listed methods for HQ and LQ image qualities.
These performance gains on both ACC and AUC scores are mainly due the the high-level and low-level joint learning strategy adopted in our formulation, further demonstrating the effectiveness of the proposed model.

\subsubsection{Detection accuracy on specific manipulation methods} Detecting low-quality manipulated face images is a challenging task as severe compression erases much detailed information from the original faces. To demonstrate that the proposed model can achieve remarkable detection performance on low-quality faces, we list the detection accuracy (ACC) of specific manipulation methods on the low-quality FF++ dataset in Table~\ref{eval_single}. DF, FF, FS, and NT represent four manipulation methods: Deepfakes, Face2Face, FaceSwap, and NeuralTextures. The manipulation-specific detectors are trained and tested on the same face manipulation methods. Comparing with previous detection methods, the proposed model achieves the best detection accuracy on all four manipulation methods.

\begin{table}
  \caption{Detection performance trained and evaluated on all manipulated and pristine data. HQ and LQ denote the high-quality and low-quality data. Compared to previous detection methods, the proposed approach achieves superior quantitative results for high-quality and low-quality face images.}
  \label{eval_all}
  \centering
  \renewcommand\arraystretch{1.15}
  \begin{tabular}{|c|c|c|c|c|}
    \hline
      &
    \multicolumn{2}{c|}{LQ (QP=40)} &
    \multicolumn{2}{c|}{HQ (QP=23)} \\
    \cline{1-5}
    Methods  & AUC(\%) & ACC(\%) & AUC(\%) & ACC(\%) \\
    \hline
    Steg. Features \cite{fridrich2012rich} & - & 55.98 & - & 70.97 \\
    \hline
    Cozzolino \textit{et al.} \cite{cozzolino2017recasting} & - & 58.69 & - & 78.45 \\
    \hline
    Bayar \& Stamm  \cite{bayar2016deep} & - & 66.84 & - & 82.97 \\
    \hline
    Rahmouni \textit{et al.} \cite{rahmouni2017distinguishing} & - & 61.18 & - & 79.08\\
    \hline
    MesoNet \cite{afchar2018mesonet} & - & 70.47 & - & 83.10\\
    \hline
    Xception \cite{chollet2017xception} & 84.38 & 83.61 & 99.71 & 98.04 \\
    \hline
    Face X-ray \cite{li2020face} & 61.60 & - & 87.35 & - \\
    \hline
    Two-branch \cite{masi2020two} & 86.59 & - & 98.70 & - \\
    \hline
    SPSL \cite{liu2021spatial} & 82.82 & 81.57 & 95.32 & 91.50\\
    \hline
    Nirkin \textit{et al.} \cite{nirkin2021deepfake} & - & 80.18 & - & -\\
    \hline
    F$^3$Net \cite{qian2020thinking}& \bm{$87.26$} & \underline{$84.65$} & 99.60 & 97.49 \\
    \hline
    Multi-Att \cite{zhao2021multi} & 85.52 & 84.49 & \underline{$99.75$} & \underline{$98.33$} \\
    \hline
    Ours & \underline{$87.10$}& \bm{$84.84$} & \bm{$99.77$} & \bm{$98.40$}\\
    \hline
\end{tabular}
\end{table}

\begin{figure*}[h]
\centering
\includegraphics[scale=0.37]{ Featuremap5.pdf}
\caption{Feature map visualization of the Xception baseline and the proposed method. }
\label{featuremap}
\end{figure*}

\subsubsection{Cross-dataset evaluation} 
Most existing detection models always suffer a significant performance drop when applied to unseen datasets. To comprehensively evaluate the generalization ability of the proposed model, we conduct extensive cross-dataset evaluation experiments in this paper. We train our model on the Deepfakes (HQ and LQ) and Pristine (HQ and LQ) data of FF++ and test it on the unseen CelebDF \cite{li2020celeb}, DFD \cite{dfd}, DFDC \cite{dolhansky2020deepfake}, and Deepfake-TIMIT \cite{korshunov2018deepfakes} datasets. For a fair comparison, we fully align the training, validation, and testing data. We list the dataset split strategy in Table~\ref{data_cross}. The training set includes 720 HQ, 720 LQ Deepfake videos, and 720 HQ, 720 LQ Pristine videos in FF++, and 270 frames are extracted from each training video. For the validation and test set (intra), we extract 100 frames for each video. For the unseen Deepfake datasets, we extract 30 frames from each video at equal intervals.

The intra- and cross-dataset evaluation results (AUC \& EER) are listed in Table~\ref{eval_crossdataset}. For all detection methods listed in Table~\ref{eval_crossdataset}, we select the checkpoint with the best AUC score on the validation set and test it on all test sets. In this part, we apply the designed two-stream learning scheme on both Xception and EfficientNet backbones, which are respectively denoted as Ours (Xcep.) and Ours (Effi.) in Table~\ref{eval_crossdataset}.
Our frameworks achieve the best AUC and EER on the FF++/DF (HQ) set and the FF++/DF (LQ) set. Under the cross-dataset evaluation setting, the proposed method outperforms all the listed models in terms of the average AUC and EER values. The cross-dataset evaluation experiment demonstrates that the proposed model is capable of achieving high generalization capability. Compared with the Xception baseline, our framework gains a 6.72\% AUC and 4.76\% EER average score improvement under the cross evaluation settings. This indicates that the proposed two-stream multi-scale learning framework can also boost the backbone model's generalization ability.

\subsubsection{Multi-class classification evaluation} Although identifying the authenticity of input faces is of great importance, specifying the manipulation method is also a non-trivial problem. However, correctly classifying the manipulation approach of the input face can reveal whether the identity or expression of the input face has been manipulated, which further unveils the potential intent of the forger.

\begin{table}
  \caption{Detection accuracy (ACC) of specific manipulation methods on low-quality faces (DF: Deepfakes, FF: Face2Face, FS: FaceSwap, NT: NeuralTextures).}
  \label{eval_single}
  \centering
  \renewcommand\arraystretch{1.15}
  \scalebox{1.1}{\begin{tabular}{|c|c|c|c|c|}
    \hline
    Methods  & DF(\%) & FF(\%) & FS(\%) & NT(\%) \\
    \hline
    Steg. Features \cite{fridrich2012rich} & 67.00 & 48.00 & 49.00 & 56.00\\
    \hline
    Cozzolino \textit{et al.} \cite{cozzolino2017recasting} & 75.00 & 56.00 & 51.00 & 62.00 \\
    \hline
    Bayar \& Stamm  \cite{bayar2016deep}  & 87.00 & 82.00 & 74.00 & 74.00\\
    \hline
    Rahmouni \textit{et al.} \cite{rahmouni2017distinguishing} & 80.00 & 62.00 & 59.00 & 59.00\\
    \hline
    MesoNet \cite{afchar2018mesonet} & 90.00 & 83.00 & 83.00 & 75.00\\
    \hline
    SPSL \cite{liu2021spatial} & 93.48 & 86.02 & 92.26 & 76.78\\
    \hline
    Xception \cite{chollet2017xception} & 97.16 & 91.02 & 96.71 & 82.88 \\
    \hline
    Ours & \bm{$97.25$} & \bm{$94.46$} & \bm{$97.13$} & \bm{$84.63$} \\
    \hline
\end{tabular}}
\end{table}

We further evaluate the proposed model on this five-way (real and four respective manipulation methods) classification task. The classification results on the FF++ low quality (QP=40) dataset are reported in Table~\ref{five_class}. We use the bold font to highlight the best results while underlining the second-best results among all listed methods. 

Compared with the result of the Xception baseline, the proposed model equipped with the semantic segmentation module (Xception(seg)) in the second last row achieves a 5.61\% recall rate improvement, going from 75.43\% {$\rightarrow$} 81.04\%. We argue that the classification performance gain mainly benefits from the supervision of the binary mask. 

The benefit of the semantic map prediction module is twofold. First, this constrains our model to focus on the manipulated region of the input face, leading the model to mine more manipulation-related information. Second, different manipulation methods tend to leave different binary mask shapes, thus providing auxiliary information to perform the five-class classification. 

As such, the scheme of binary mask supervision can  separate the faces manipulated by different approaches. To better clarify this point, we further present the visualization of predicted mask maps on the testing set in Fig.~\ref{Biasmap}. The mean map on the left represents the average summation of overall predicted localization maps and pristine maps. The top row shows the mean maps of four specific manipulation methods and pristine face images. The absolute difference maps between the top-row maps and the mean map are shown in the bottom row. It can be readily observed that each bias map is discriminative from the others.

The final classification result in the last row of Table~\ref{five_class} shows that the proposed model outperforms the previous state-of-the-art method, SPSL, by 3.25\% average recall rate, going from 78.94\% {$\rightarrow$} 82.19\%. The noise map prediction module also mines significant manipulation method clues. Thus, we can observe that the final result (Xception(fusion)) achieves a 1.15\% average recall rate and a 6.22\% pristine recall rate improvement comparing with the results of Xception(seg), demonstrating the effectiveness of the noise map prediction module from a complementary viewpoint. 

Furthermore, we show the t-SNE feature embedding visualization of the Xception baseline and the proposed method in Fig.~\ref{tsne}. As shown in Fig.~\ref{tsne} (a), the Xception model cannot separate different manipulation methods and pristine face images, and the pristine faces are clustered with NeuralTextures and Face2Face fake faces in the feature space. However, the proposed model achieves a better multi-class embedding division performance in the t-SNE feature space, and the pristine data is less confused with other manipulated data, demonstrating the effectiveness of the proposed model from another point of view.

\subsubsection{Feature map visualization} To better demonstrate the effectiveness of the proposed model, we further visualize the feature maps of the Xception baseline and the proposed model ones in Fig.~\ref{featuremap}. All real and fake faces are from the FF++ low-quality dataset, and the models are trained under the setting of Sec.IV-C.1). 

As shown in Fig.~\ref{featuremap}, the baseline feature maps of real and corresponding fake faces are similar, resulting in a struggling manipulation detection performance. Conversely, the proposed model prefers to focus on the central regions for fake faces and the peripheral regions for pristine faces. As a result, comparing with the Xception baseline, the real and fake feature maps extracted from the proposed model are more discriminative, thus leading to a better manipulation detection performance.

\begin{figure*}[h]
\centering
\includegraphics[scale=0.58]{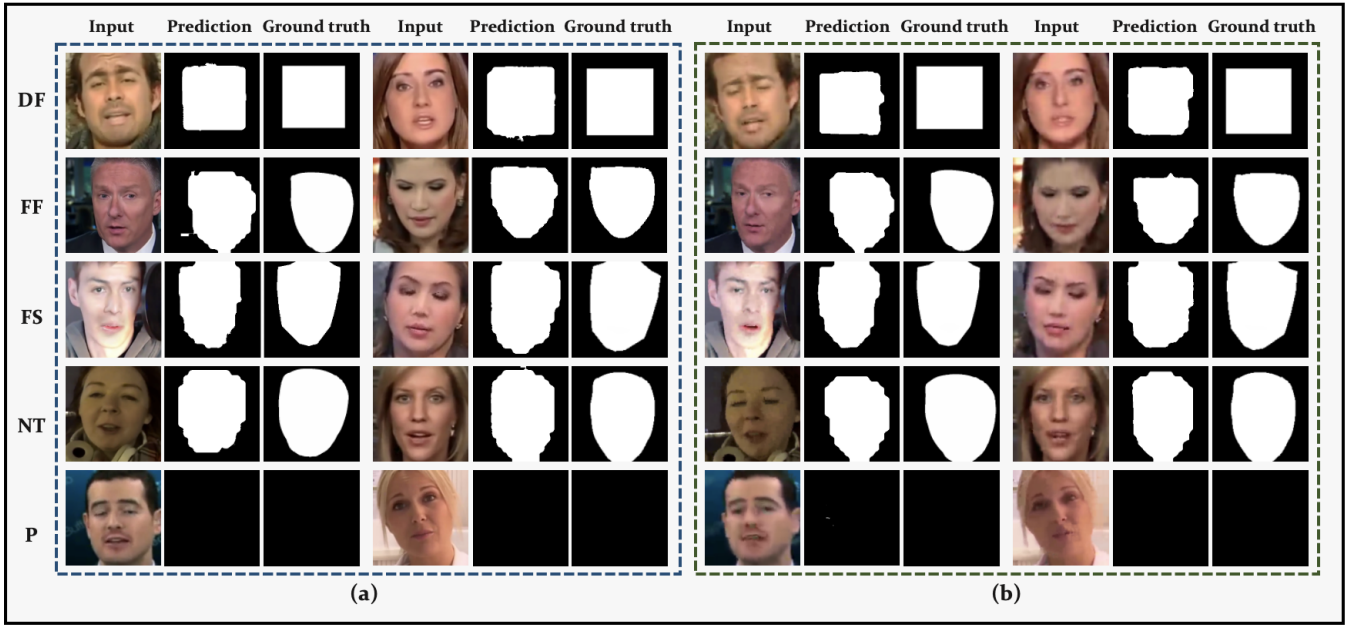}
\caption{Face manipulation localization results on (a) HQ faces (QP=23) and (b) LQ faces (QP=40). DF: Deepfakes, FF: Face2Face, FS: FaceSwap, NT: NeuralTextures, P: Pristine. The input face, predicted forgery localization map, and corresponding ground truth mask are presented in the left, middle, and right columns. It can be observed that the proposed method can locate the manipulated regions of the input faces with different head poses and various lighting conditions.}
\label{Seg}
\end{figure*}

\begin{figure*}[h]
\centering
\includegraphics[scale=0.24]{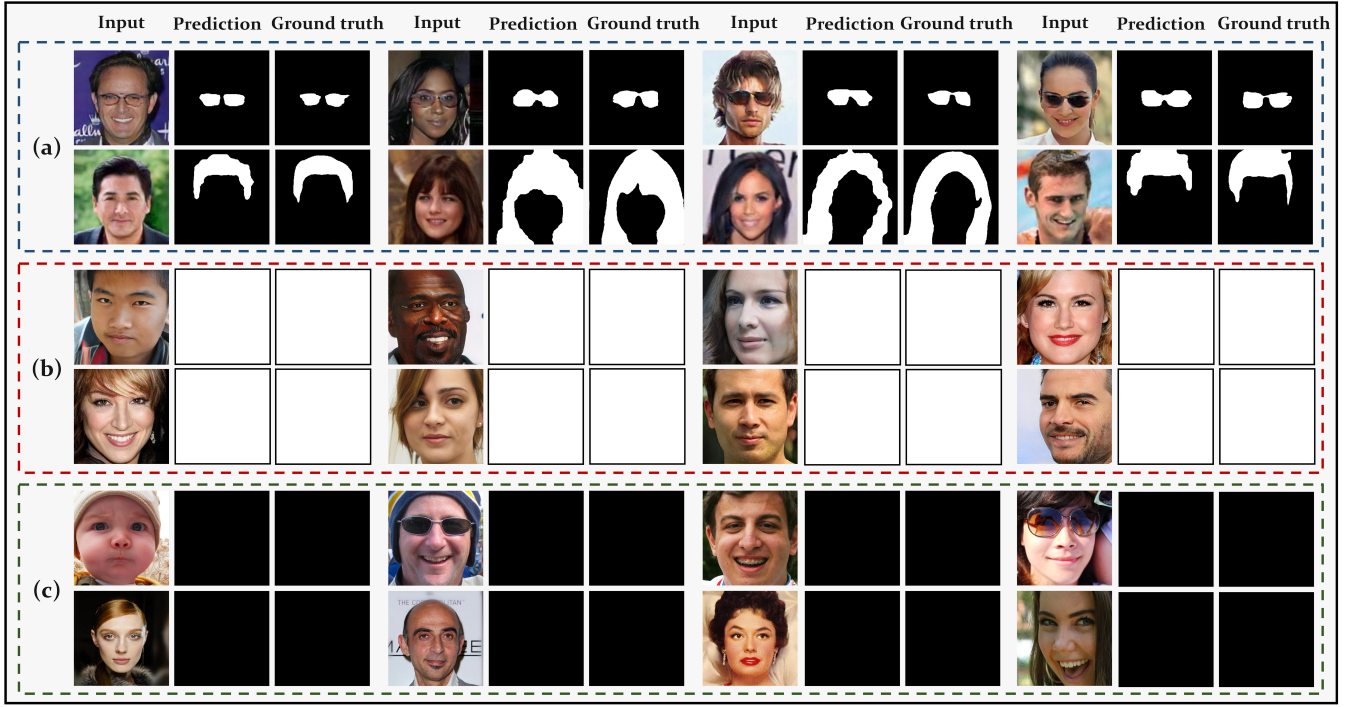}
\caption{Face manipulation localization results on our collected dataset. (a). attribute manipulated faces; (b). entirely synthetic faces; (c). real faces. }
\label{seg_kcq}
\end{figure*}

\begin{table}
  \caption{Dataset Breakdown for training, validation, and testing sets under the cross-dataset evaluation setting.}
  \label{data_cross}
  \centering
  \renewcommand\arraystretch{1.15}
  \scalebox{0.88}{\begin{tabular}{|c|c|c|c|c|c|}
    \hline
    \multicolumn{2}{|c|}{} &
    \multicolumn{2}{c|}{\textbf{\# Video}}  &
    \multicolumn{2}{c|}{\textbf{\# Frame}} \\
    \hline
     & \textbf{Dataset} & Real & Fake & Real & Fake \\
    \hline
     \textbf{Training} & FF++ & 1,440 & 1,440 & 388,691 & 386,323 \\
    \hline
    \textbf{Validation} & FF++ & 280 & 280 & 28,000 & 27,818 \\
    \hline
    \textbf{Test (Intra)} & FF++ & 280 & 280 & 28,000 & 27,957 \\
    \hline
    \multirow{6}*{\textbf{Test (Cross)}} & CelebDF & 590 & 5,639 & 17,700 & 169,170 \\
	\cline{2-6}
	~ & DFD (HQ) & 363 & 3,068 & 10,890 & 92,040 \\
	\cline{2-6}
    ~ & DFD (LQ) & 363 & 3,068 & 10,890 & 92,040 \\
    \cline{2-6}
    ~ & DFDC & 19,154 & 100,000  & 574,620 & 3,000,000 \\
    \cline{2-6}
    ~ & DF-TIMIT (HQ) & 240 & 320 & 7,200 & 9,600 \\
    \cline{2-6}
    ~ & DF-TIMIT (LQ) & 240 & 320 & 7,200 & 9,600 \\
    \hline
\end{tabular}}
\end{table}

\begin{table*}
  \caption{Cross-dataset evaluation results.}
  \label{eval_crossdataset}
  \centering
  \renewcommand\arraystretch{1.15}
  \scalebox{0.82}{\begin{tabular}{|c|c|c|c|c|c|c|c|c|c|c|c|c|c|c|c|c|c|c|}
    \hline
    &\multicolumn{4}{c|}{\textbf{INTRA (\%)}} &
    \multicolumn{14}{c|}{\textbf{CROSS (\%)}} \\
    \hline
     \textbf{Dataset} & \multicolumn{2}{c|}{FF++/DF (HQ)} & \multicolumn{2}{c|}{FF++/DF (LQ)} & \multicolumn{2}{c|}{CelebDF}  & \multicolumn{2}{c|}{DFD (HQ)} & \multicolumn{2}{c|}{DFD (LQ)} & \multicolumn{2}{c|}{DFDC} & \multicolumn{2}{c|}{DF-TIMIT (HQ)} & \multicolumn{2}{c|}{DF-TIMIT (LQ)} & \multicolumn{2}{c|}{\textbf{AVG (Cross)}}\\
    \hline
     \textbf{Method} & AUC & EER & AUC & EER & AUC & EER & AUC & EER & AUC & EER & AUC & EER & AUC & EER & AUC & EER & AUC & EER \\
    \hline
    MesoNet \cite{afchar2018mesonet} & 76.34 & 30.23 & 70.11 & 35.39 & 58.85 & 43.89 & 62.07 & 41.75 & 52.25 & 48.65 & 54.60 & 46.67 & 33.61 & 60.16 & 45.08 & 53.04 & 51.08 & 49.03 \\
    \hline
    MesoIncep4 \cite{afchar2018mesonet} & 98.35 & 6.72 & 95.56 & 11.68 & 68.26 & 37.59 & 79.18 & 27.68 & 63.27 & 40.37 & 61.92 & 41.12 & 16.12 & 76.18 & 27.47 & 66.77 & 52.70 & 48.29\\
    \hline
    ResNet50 \cite{he2016deep} & 99.15 & 4.35 & 98.46 & 6.57 & 67.09 & 37.57 & 69.60 & 35.42 & 60.61 & 42.23 & 61.97 & 41.19 & 41.95 & 55.97 & 47.27 & 52.33 & 58.08 & 44.12 \\
    \hline
    Face X-ray \cite{li2020face} & 99.60 & 2.45 & 99.06 & 4.41 & 71.89 & 32.64 & 69.61 & 33.97 & 62.89 & 39.58 & 58.97  & 43.52 & 42.52 & 55.07 & 50.05 & 49.11  & 59.32 & 42.31 \\
    \hline
    DFFD \cite{dang2020detection} & 99.74 & 2.56 & 99.37 & 3.92 & 69.55 & 35.92 & 71.69 & 33.76 & 60.60 & 42.32 & 59.72 & 42.92 & 32.91 & 61.16 & 39.32 & 57.06 & 55.63 & 45.52 \\
    \hline
    Multi-task \cite{nguyen2019multi} & 98.57 & 5.94 & 96.02 & 10.62 & 65.18 & 38.92 & 70.75 & 34.55 & 58.61 & 44.49 & 57.38 & 44.83 & 16.53 & 77.86 & 15.59 & 78.50 & 47.34 & 53.19 \\
    \hline
    EfficientNet \cite{tan2019efficientnet} & 99.65 & 1.93 & 99.19 & 4.06 & 75.90 & 30.94 & 80.63 & 27.07 & 64.19 & 39.87 & 66.39 & 38.68 & 29.12 & 65.75 & 28.34 & 65.25 & 57.43 & 44.59 \\
    \hline
    F$^3$Net \cite{qian2020thinking} & 99.74 & 2.16 & 99.21 & 4.47 & 72.28 & 33.85 & 72.92 & 33.04 & 58.89 & 43.82 & 63.33 & 40.54 & 38.55 & 58.33 & 45.67 & 52.72 & 58.61 & 43.72 \\
    \hline
    Xception \cite{chollet2017xception} & 99.64 & 2.78 & 99.16 & 4.23 & 67.75 & 37.26 & 72.45 & 33.50 & 59.73 & 43.12 & 63.12 & 40.58 & 33.82 & 62.83 & 40.79 & 57.44 & 56.28 & 45.79 \\
    \hline
    \hline
    Ours (Effi.) & 99.73 & 1.67 & \textbf{99.46} & \textbf{2.92} & 67.15 & 37.93 & 73.52 & 33.06 & 67.21 & 37.37 & 60.32 & 43.12  & 49.90 & 49.47 & 51.01 & 48.03 & 61.52& 41.50\\
    \hline
    Ours (Xcep.) & \textbf{99.85} & \textbf{1.80} & 99.30 & 4.10 & 70.65 & 35.45 & 76.23 & 30.26 & 64.53 & 39.59 & 63.31 & 40.79 & 47.20 & 53.23  & 56.08 & 46.86 & \textbf{63.00} & \textbf{41.03} \\
    \hline
\end{tabular}}
\end{table*}

\subsubsection{Discussions}
In this subsection, we quantitatively report the face forgery detection performances on various experimental settings and qualitatively visualize the feature maps and t-SNE feature embedding distributions. The presented results demonstrate the effectiveness and robustness of the proposed framework, which can capture the artifacts in both semantic-level and noise-level, thus achieving superior detection performance on all mentioned experimental settings.

\begin{table}
  \caption{Recall rates of genuine and four manipulation methods with c40(QP=40) setting in the five-class classification. DF: Deepfakes, FF: Face2Face, FS: FaceSwap, NT: NeuralTextures, P: Pristine. We use the bold font to highlight the best results while underlining the second-best results among all listed methods.}
  \label{five_class}
  \centering
  \renewcommand\arraystretch{1.15}
  \scalebox{0.82}{\begin{tabular}{|c|c|c|c|c|c|c|}
    \hline
    Methods  & DF(\%) & FF(\%) & FS(\%) & NT(\%) & P(\%) & AVG(\%) \\
    \hline
    MesoNet \cite{afchar2018mesonet} & 62.45 & 40.37 & 28.89 & 63.35 & 40.93 & 47.20 \\
    \hline
    ResNet50 \cite{he2016deep} & 82.76 & 76.30 & 81.21 & 52.18 & 46.82& 67.85 \\
    \hline
    Xception \cite{chollet2017xception} & 86.61 & 78.88 & 83.16 & 52.94 & \underline{$75.55$} & 75.43 \\
    \hline
    SPSL \cite{liu2021spatial} & 91.16 & 78.31 & 88.75 & 58.97 & \bm{$77.49$} & 78.94 \\
    \hline
    Xception(seg) (Ours)  & \underline{$95.15$} & \bm{$87.61$} & \underline{$92.08$} & \bm{$72.82$} & 57.58 & \underline{$81.04$}\\
    \hline
    Xception(fusion) (Ours) & \bm{$95.28$} & \underline{$86.96$} & \bm{$93.24$} & \underline{$71.66$} & 63.80 & \bm{$82.19$}\\
    \hline
    
\end{tabular}}
\end{table}

\subsection{Manipulation Localization}

In this subsection, we evaluate the manipulation localization performance of the proposed model both quantitatively and qualitatively. 

\subsubsection{Quantitative evaluation} As we apply the multi-scale feature learning strategy in the proposed framework, three predicted manipulation maps can be obtained for each given face. Herein, we apply a weighted summation strategy over the three predicted maps to determine the final manipulation localization map:
\begin{equation}
     \hat{M_{f}} = {\gamma}_{1} \hat{M}_{1}^{\uparrow} + {\gamma}_{2} \hat{M}_{2}^{\uparrow} + {\gamma}_{3} \hat{M}_{3}^{\uparrow},
\end{equation}
where {$\hat{M}_{f}$} represents the final determined manipulation localization map. The parameters {${\gamma}_{1}$}, {${\gamma}_{2}$}, and {${\gamma}_{3}$} are used to weigh the predicted maps. We resize the predicted masks {$\hat{M}_{1}$}, {$\hat{M}_{2}$}, and {$\hat{M}_{3}$} ({$\hat{M}_{1}$}, {$\hat{M}_{2}$}, and {$\hat{M}_{3}$} are specified in Eq. (3)) to the original input face size 299 {$\times$} 299, obtaining the aligned manipulation localization maps {$\hat{M}_{1}^{\uparrow}$}, {$\hat{M}_{2}^{\uparrow}$}, and {$\hat{M}_{3}^{\uparrow}$}.

Intuitively, the features extracted from the deep layer tend to carry rich semantic information. Thus we allocate relatively larger weight to {$\hat{M}_{3}^{\uparrow}$}. To study the best weighting strategy, we report the Intersection over Union (IoU) score for both high-quality and low-quality faces in Table~\ref{eval_hyper3}. Comparing the IoUs in the first row ({${\gamma}_{1}$}=0.0, {${\gamma}_{2}$}=0.0, {${\gamma}_{3}$}=1.0) and the IoUs in the fourth row ({${\gamma}_{1}$}=0.1, {${\gamma}_{2}$}=0.2, {${\gamma}_{3}$}=0.7), we can conclude that the multi-scale feature learning strategy benefits the manipulation localization performance because leveraging the information of all three features leads to a better localization accuracy. As such, the values of {${\gamma}_{1}$}, {${\gamma}_{2}$}, and {${\gamma}_{3}$} are set as 0.1, 0.2, and 0.7. Our model can achieve remarkable 0.8413 and 0.9305 IoU localization performance for low-quality and high-quality faces, respectively, demonstrating the effectiveness of the designed semantic map prediction module.

\begin{table}
  \caption{Face manipulation localization accuracy (IoU) for low-quality (LQ) and high-quality (HQ) faces with different weighting strategies.}
  \label{eval_hyper3}
  \centering
  \renewcommand\arraystretch{1.15}
  \scalebox{1.25}{\begin{tabular}{|c|c|c|c|c|}
    \hline
    {$\gamma_1$}  & {$\gamma_2$} & {$\gamma_3$} & IoU (LQ) & IoU (HQ)  \\
    \hline
    0.0  & 0.0 & 1.0 & 0.8399 & 0.9277 \\
    \hline
    0.0  & 0.1 & 0.9 & 0.8408 & 0.9291 \\
    \hline
    0.1  & 0.1 & 0.8 & 0.8406 & 0.9304 \\
    \hline
    0.1  & 0.2 & 0.7 & \bm{$0.8413$}  & \bm{$0.9305$}\\
    \hline
    0.1  & 0.3 & 0.6 & 0.8371 & 0.9240\\
    \hline
    0.1  & 0.4 & 0.5 & 0.8270 & 0.9110\\
    \hline
    0.2  & 0.2 & 0.6 & 0.8373 & 0.9247\\
    \hline
    0.2  & 0.3 & 0.5 & 0.8313 & 0.9177\\
    \hline
    0.2  & 0.4 & 0.4 & 0.8205 & 0.9015\\
    \hline
\end{tabular}}
\end{table}

\subsubsection{Qualitative evaluation on FF++ dataset} To further study the manipulation localization performance of the proposed model, we qualitatively present the results of face manipulation localization on high-quality faces and low-quality faces in Fig.~\ref{Seg}  (a) and Fig.~\ref{Seg} (b). The faces in different rows represent the fake faces created by corresponding manipulation methods and pristine real faces. The input face, predicted localization map and the corresponding ground truth mask are shown in the left, middle, and right columns. Comparing with the ground-truth masks, it can be seen that the manipulated regions can be well captured for both low-quality and high-quality faces by the proposed model. Furthermore, our model can  accurately determine the forgery locations for the faces with various head-poses and very poor lighting conditions  (e.g., NT 1st example), which further validates the robustness of the proposed model.

\subsubsection{Qualitative evaluation on our collected dataset} In addition to the FF++ dataset, which manipulates faces in the whole face region, we further consider two pervasive face manipulation types in this subsection: entirely synthetic faces and attribute manipulated faces. Herein, we employ the popular StyleGAN \cite{karras2019style} and PGGAN \cite{karras2018progressive} to produce 40000 entirely synthetic faces and use StarGAN \cite{choi2018stargan} and AttGAN \cite{he2019attgan} to generate 40000 attribute-manipulated faces (two attribute manipulations are considered in this paper: glasses and hairs). We further collect 40000 real faces from the FFHQ \cite{karras2019style} and CelebA-HQ \cite{karras2018progressive} datasets. The training, validation, and test sets are split as 8:1:1. We denote the collected Face Manipulation Localization Dataset as FMLD. The manipulation localization performance of the proposed model on FMLD is shown in Fig.~\ref{seg_kcq}. Again, the proposed model achieves outstanding forgery localization performance.

\subsubsection{Comparison with SOTA forgery localization methods} 
\begin{figure*}[h]
\centering
\includegraphics[scale=0.50]{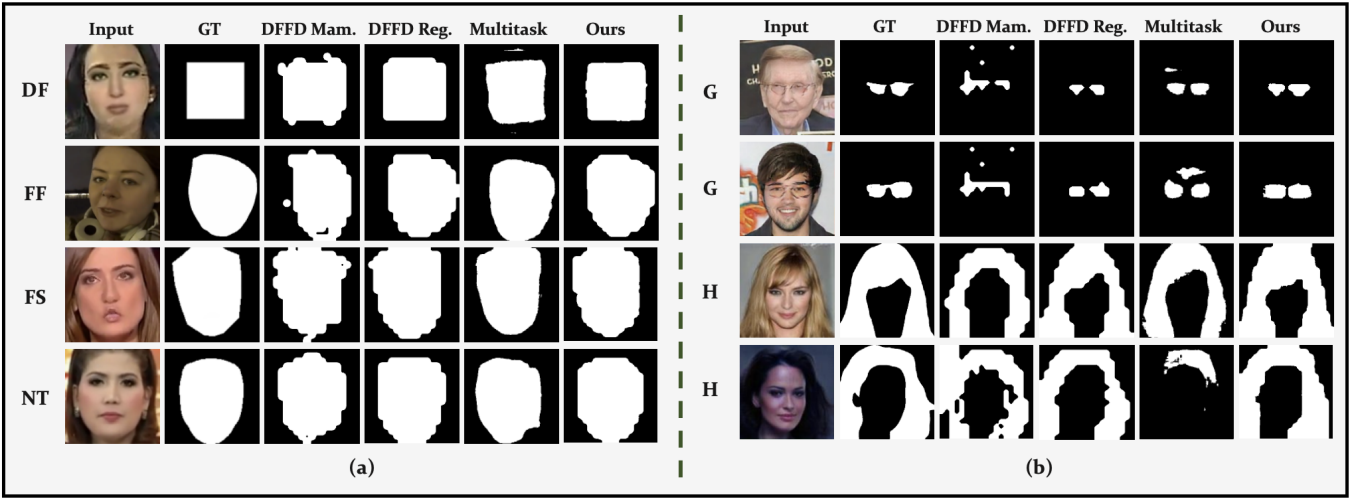}
\caption{Face manipulation localization results of different methods on (a) FF++ HQ dataset and (b) FMLD. DF, FF, FS, and NT represent Deepfakes, Face2Face, FaceSwap, and NeuralTextures. G and H indicate two attribute manipulations: glasses and hair. }
\label{seg_sample}
\end{figure*}

In this part, we compare the proposed model with other methods using masks in the training stage. We respectively report the manipulation detection and localization results in Table~\ref{mask_method} and Table~\ref{loc_quan}. It is worth mentioning that noise information is not used in our method in these two evaluations for fair comparisons. We present the detection performance on FF++ LQ (QP=40) dataset in Table~\ref{mask_method}. Multitask \cite{nguyen2019multi} designs a U-shape CNN that simultaneously detects manipulated faces and locates the manipulated regions for each query. DFFD \cite{dang2020detection} uses mask attention mechanism to perform face manipulation detection and localization, and DFFD Reg. and DFFD Mam. are two different attention mechanisms of DFFD. In this table, we use the bold font to highlight the best results while underlining the second-best results among all listed methods. It can be observed that our method achieves the best AUC, ACC, AP, and EER scores and the second-best FPR and FNR performance. These performance boosts  benefit from the proposed multi-scale mask supervision. Moreover, we further report the manipulation localization results on the FF++ HQ (QP=23) and FMLD datasets of each method in Table~\ref{loc_quan}. Following DFFD, we use Pixel-wise Binary Classification Accuracy (PBCA) and Inverse Intersection Non-Containment (IINC) as the evaluation metrics. In Table~\ref{loc_quan}, the proposed method achieves the best localization performance on both FF++ HQ and FMLD datasets. We further present localization examples of these methods in Fig.~\ref{seg_sample}. Compared with Multitask, our method has a superior pixel-wise localization accuracy. Compared with DFFD, the proposed method has a more precise localization performance on the boundary regions, which mainly benefit from our multi-scale learning strategy.

\begin{table}
  \caption{Detection performance on FF++ LQ (QP=40) dataset compared with the methods using masks.}
  \label{mask_method}
  \centering
  \renewcommand\arraystretch{1.15}
  \scalebox{0.75}{\begin{tabular}{|c|c|c|c|c|c|c|}
    \hline
    Methods  & AUC(\%) $\uparrow$ & ACC(\%) $\uparrow$ & AP(\%) $\uparrow$ & FPR(\%) $\downarrow$ & FNR(\%) $\downarrow$ & EER(\%) $\downarrow$ \\
    \hline
    Multitask & 75.73 & 75.10 & 91.76 & \bm{$43.77$} & 20.18 & 30.79 \\
    \hline
    DFFD Reg. & \underline{$84.95$} & \underline{$83.70$} & \underline{$95.24$} & 48.78 & \bm{$8.18$} & \underline{$23.03$} \\
    \hline
    DFFD Mam. & 83.87 & 83.63 & 94.73 & 46.06 & 8.95 & 23.32 \\
    \hline
    Ours & \bm{$86.87$} & \bm{$84.56$} & \bm{$95.98$} & \underline{$44.18$} & \underline{$8.26$} & \bm{$21.03$} \\
    \hline
\end{tabular}}
\end{table}

\begin{table}
  \caption{Manipulation localization performance compared with state-of-the-arts localization methods.}
  \label{loc_quan}
  \centering
  \renewcommand\arraystretch{1.15}
  \scalebox{0.9}{\begin{tabular}{|c|c|c|c|c|}
    \hline
     Dataset &
    \multicolumn{2}{c|}{FF++ HQ} &
    \multicolumn{2}{c|}{FMLD} \\
    \cline{1-5}
    Methods  & PBCA(\%) $\uparrow$ & IINC(\%) $\downarrow$ & PBCA(\%) $\uparrow$ & IINC(\%) $\downarrow$ \\
    \hline
    Multitask \cite{nguyen2019multi} & 94.88 & 4.46 & 98.59 & 3.52\\
    \hline
    DFFD Reg. \cite{dang2020detection} & 94.85 & 4.57 & 98.72 & 3.31\\
    \hline
    DFFD Mam. \cite{dang2020detection} & 91.45 & 13.09 & 96.86 & 23.93\\
    \hline
    Ours & \bm{$95.77$} & \bm{$3.62$} & \bm{$99.06$} & \bm{$2.53$} \\
    \hline
\end{tabular}}
\end{table}

\subsubsection{Discussion}

In this subsection, we quantitatively and qualitatively evaluate the manipulation localization performance. The proposed model captures the high-level semantic information from the multi-scale features, with state-of-the-art localization accuracy and segmentation performance for FF++ faces, attribute-manipulated faces, and entirely synthetic faces.

\subsection{Ablation Study}
To validate the effectiveness of the semantic map prediction module, the noise map prediction module, and the multi-scale feature extraction strategy in the proposed framework, we conduct extensive ablation experiments in this subsection.

\subsubsection{Effectiveness of sub-modules} 
To study the effectiveness of the semantic map prediction module and the noise map prediction module, we report the manipulation detection results of the Xception backbone equipped with different modules on low-quality faces in Table~\ref{eval_abl1}. By comparing with the AUC, AP and EER scores in the third row and the detection result of the Xception baseline, we can conclude that the semantic map prediction module leads the model to mine more artifact clues and improve the detection performance. On the other hand, the detection performance drops slightly when equipping the noise prediction module on the Xception backbone. The reason is that forcing the model to only focus on the noise-level clues may cause severe image information loss, resulting in detection performance drops. In this work, the noise clue is regarded as the auxiliary information that plays a complementary role to the semantic map prediction module. 

\begin{figure}[h]
\centering
\includegraphics[scale=0.28]{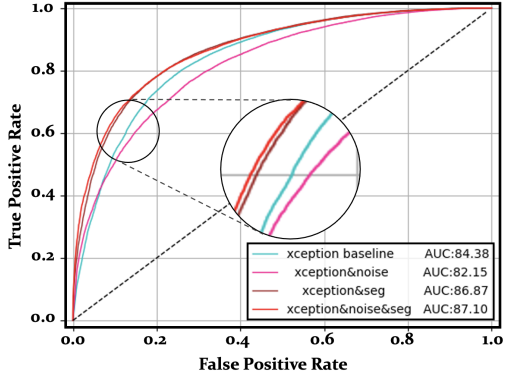}
\vspace{-0.35cm}
\caption{Results for the Xception baseline and the three proposed models (ROC).}
\label{roc_module}
\end{figure}

Comparing with the detection results in the third row, the detection performance in the last row gains specific improvement, proving the effectiveness of the noise map prediction module. To better validate the effectiveness of each module, we also show the ROC curves in Fig.~\ref{roc_module}. We can observe that the proposed model achieves the best AUC performance and the best TPR at a lower FPR value, while low FPR is one of the most challenging face manipulation detection scenarios \cite{qian2020thinking}.

We can be concluded that the semantic map prediction module and the noise map prediction module indeed help improve the detection capability of the proposed model.

\subsubsection{Effectiveness of multi-scale feature learning} 
We further conduct an ablation study to demonstrate the effectiveness of the multi-scale feature learning strategy. The manipulation detection performance of different feature extraction strategies is listed in Table~\ref{eval_abl2}. As illustrated in Fig.~\ref{overview}, {$f_{1}$}, {$f_{2}$}, and {$f_{3}$} represent the features extracted from the shallow, middle, and deep layers of the Xception backbone, and each feature layer will be processed by the semantic map prediction module and the noise map prediction module. 

\begin{figure}[h]
\centering
\includegraphics[scale=0.28]{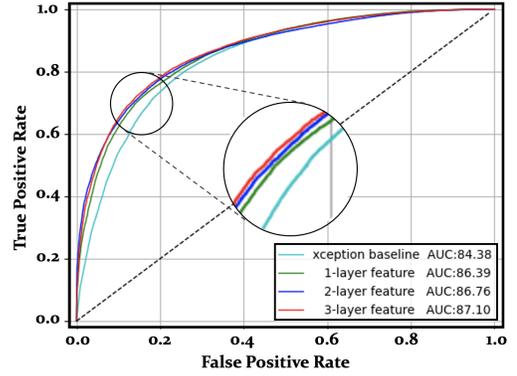}
\vspace{-0.35cm}
\caption{Results for the Xception baseline and the proposed models with three different feature extraction strategies (ROC).}
\label{roc_feature}
\end{figure}

To better illustrate the effectiveness of each feature layer, we also show the ROC curves of different feature extraction strategies in Fig.~\ref{roc_feature}. 
From the results in Table~\ref{eval_abl2} and Fig.~\ref{roc_feature}, we can conclude that 1) the usage of deep feature layer {$f_{3}$} leads to the most AUC score gain as it contains much semantic information; and 2) each extracted feature layer contributes the final detection performance improvements, verifying the effectiveness of the proposed multi-scale feature learning strategy.

\begin{table}
  \caption{AUC scores of the Xception backbone equipped with semantic map prediction module and noise map prediction module on low-quality faces.}
  \label{eval_abl1}
  \centering
  \renewcommand\arraystretch{1.15}
  \scalebox{1.0}{\begin{tabular}{|c|c|c|c|c|c|}
    \hline
    Xception & Noise & Mask & AUC(\%) $\uparrow$ & AP(\%) $\uparrow$ & EER(\%) $\downarrow$\\
    \hline
    $\surd$ & - & -  & 84.38 & 94.63 & 22.97\\
    \hline
    $\surd$ & $\surd$ & -  & 82.15 & 94.40 & 24.84\\
    \hline
    $\surd$ & - & $\surd$  & 86.87 & 95.98 & 21.03 \\
    \hline
    $\surd$ & $\surd$ & $\surd$  & \bm{$87.10$}  & \bm{$96.18$} & \bm{$20.94$}\\
    \hline
\end{tabular}}
\end{table}

\begin{table}
  \caption{Ablation study of the proposed model with different feature layer extraction strategies on low-quality faces. }
  \label{eval_abl2}
  \centering
  \renewcommand\arraystretch{1.15}
  \scalebox{1.0}{\begin{tabular}{|c|c|c|c|c|c|c|}
    \hline
    Xception &{$f_{3}$} & {$f_{2}$} & {$f_{1}$} & AUC(\%) $\uparrow$ & AP(\%) $\uparrow$ & EER(\%) $\downarrow$ \\
    \hline
    $\surd$ & - & - & - & 84.38 & 94.63 & 22.97\\
    \hline
    $\surd$ & $\surd$ & - & - & 86.39 & 95.87 & 22.02\\
    \hline
    $\surd$ & $\surd$ & $\surd$ & - &  86.76 & \bm{$96.20$} & 21.27\\
    \hline
    $\surd$ & $\surd$ & $\surd$ & $\surd$ & \bm{$87.10$} & 96.18 & \bm{$20.94$}\\
    \hline
\end{tabular}}
\end{table}

\subsubsection{Study of feature aggregation strategy} 
\begin{figure*}[h]
\centering
\includegraphics[scale=0.5]{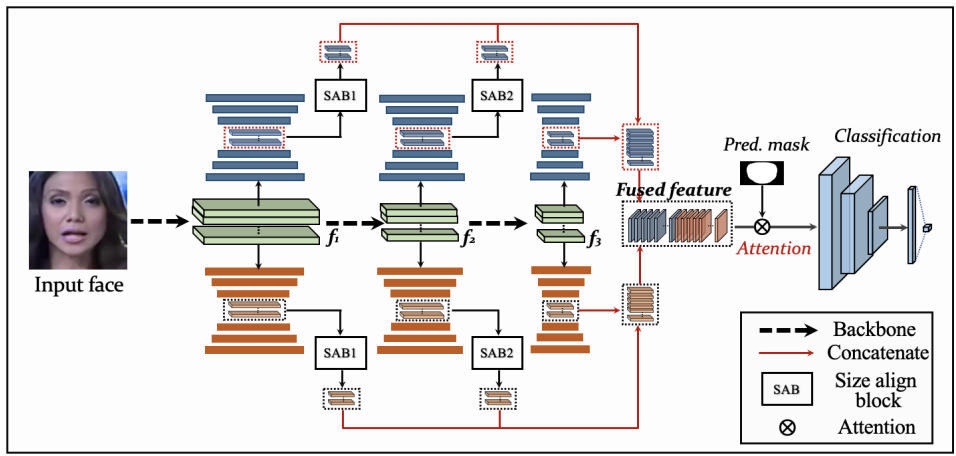}
\vspace{-0.35cm}
\caption{Illustration of feature aggregation strategy. The $f_{1}$, $f_{2}$, and $f_{3}$ are synchronously fed to semantic map prediction module and noise map prediction module. Sequentially, we feed forward $f_{1}$ and $f_{2}$ into Size Align Blocks (SAB) that consist of several convolution and pooling layers for feature size alignment. The aligned noise and mask features are concatenated along the channel dimension. Then, we conduct the feature spatial attention by multiplying the mask map predicted by the last $f_{3}$-based semantic map prediction module with the fused feature. And the attended feature is finally sent to the classification module to identify the authenticity of input query face.}
\label{dense}
\end{figure*}

Thanks to the proposed multi-scale learning strategy, the three features extracted from the backbone should contain abundant information. In this part, we further study the impact of the feature aggregation on the forgery classification performance. The illustration of the feature aggregation strategy is shown in Fig.~\ref{dense}. The $f_{1}$, $f_{2}$, and $f_{3}$ are synchronously fed to semantic map prediction module and noise map prediction module. Sequentially, we feed forward $f_{1}$ and $f_{2}$ into Size Align Blocks (SAB) that consist of several convolution and pooling layers for feature size alignment. The aligned noise and mask features are concatenated along the channel dimension. Then, we conduct the feature spatial attention by multiplying the mask map predicted by the last $f_{3}$-based semantic map prediction module with the fused feature. The attended feature is finally sent to the classification module to identify the authenticity of the input query face. We summarize the feature aggregation process as follows, 

\begin{align}
    F_{att} = 
    &\{\underbrace{[SAB_{1}^{m}(f_{1}^{m})\oplus  SAB_{2}^{m}(f_{2}^{m}) \oplus f_{3}^{m}]}_{F_{cat}^{m}} \nonumber
    \\ &\oplus\underbrace{[SAB_{1}^{n}(f_{1}^{n})\oplus SAB_{2}^{n}(f_{2}^{n})\oplus f_{3}^{n}]}_{F_{cat}^{n}}\}\otimes M_{3}
\end{align}

where the superscripts $m$ and $n$ indicate the mask and noise prediction modules. $\oplus$ indicates the concatenation operation. $M_{3}$ is the predicted mask from the semantic map prediction module, and $\otimes$ denotes the spatial attention process.

We report the detection results of such feature aggregation strategy in Table~\ref{aggregation} (denoted as Aggregation), and the results of using single feature $f_{3}$ (denoted as Single $f_{3}$) are also listed for reference. We can observe that the feature aggregation indeed improves the detection performance. It is worth mentioning that this also makes the model bigger and requires more computational resources. 

\subsubsection{Evaluation of the proposed learning scheme on other backbones} Extensive experiments have demonstrated the effectiveness of the proposed framework on the Xception backbone. In this part, we further apply the proposed two-stream multi-scale learning architecture on the VGG16 \cite{simonyan2014very} and EfficientNet \cite{tan2019efficientnet} backbones to verify that our framework can be conveniently transferred to other detectors in a plug-and-play manner. The detection results are reported in Table~\ref{vgg16}. Unsurprisingly, the semantic-level supervision boosts the detection accuracy. And the noise fingerprint playing a complementary role further empowers the final classification performance.

\begin{table}
  \caption{Manipulation detection results compared with feature aggregation strategy.}
  \label{aggregation}
  \centering
  \renewcommand\arraystretch{1.15}
  \scalebox{1.1}{\begin{tabular}{|c|c|c|c|c|}
    \hline
    Method  & AUC(\%) $\uparrow$ & ACC(\%) $\uparrow$ & AP(\%) $\uparrow$ & EER(\%) $\downarrow$ \\
    \hline
    Aggregation  & \bm{$87.36$} & \bm{$85.02$} & \bm{$96.24$} & \bm{$20.75$} \\
    \hline
    Single $f_{3}$ & 87.10 & 84.84 & 96.18 & 20.94 \\
    \hline
\end{tabular}}
\end{table}

\begin{table}
  \caption{Detection performance of the proposed two-stream multi-scale framework on VGG16 and EfficientNet backbone.}
  \label{vgg16}
  \centering
  \renewcommand\arraystretch{1.15}
  \scalebox{0.95}{\begin{tabular}{|c|c|c|c|c|}
    \hline
    Methods  & AUC(\%) $\uparrow$ & ACC(\%) $\uparrow$ & AP(\%) $\uparrow$ & EER(\%) $\downarrow$ \\
    \hline
    \hline
    VGG16 \cite{simonyan2014very} & 81.45 & 82.46 & 93.41 & 25.49 \\
    \hline
    VGG16 \& Mask. & 82.70 & 82.53 & 94.61 & 24.99 \\
    \hline
    VGG16 \& Noise & 79.14 & 79.78 & 92.84 & 27.78 \\
    \hline
    VGG16 \& Fusion &  \bm{$83.45$} & \bm{$83.27$} & \bm{$94.62$} & \bm{$24.25$} \\
    \hline
    \hline
    EfficientNet & 84.19 & 84.05 & 94.30 & 22.88\\
    \hline
    EfficientNet \& Mask. & 87.25 & 84.70 & 96.37 & 20.64\\
    \hline
    EfficientNet \& Noise & 84.08 & 83.97 & 94.01 & 23.02\\
    \hline
    EfficientNet \& Fusion & \textbf{87.96} & \textbf{85.28} & \textbf{96.39} & \textbf{19.63} \\
    \hline
\end{tabular}}
\end{table}

\begin{figure}[h]
\centering
\includegraphics[scale=0.40]{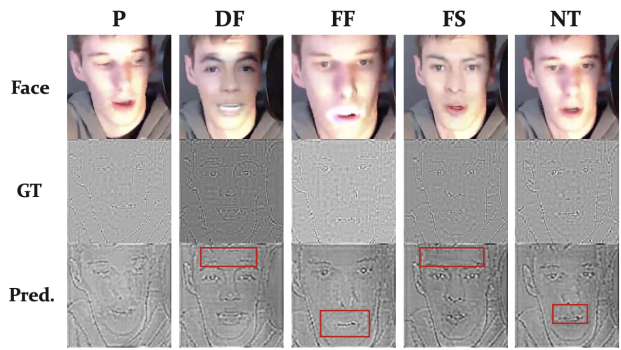}
\vspace{-0.35cm}
\caption{Visualization of the noise map predictions. The top row shows the input faces. The middle and bottom rows respectively present the ground-truth and predicted noise maps (The noise map intensity has been enlarged for better visualization). P: Pristine, DF: Deepfakes, FF: Face2Face, FS: FaceSwap, NT: NeuralTextures.}
\label{Noise_pred}
\end{figure}

\subsubsection{Effectiveness of noise features}
In the training and test stages, we exploit the noise feature rather than directly use the noise map to conduct the binary classification. Directly feeding the noise maps into the model will primarily suppress the image content and may lead the model to make wrong decisions, as noise is a weak signal and can be easily overwhelmed by imperfectly removed image content \cite{verdoliva2020media}. This work exploits noise fingerprints more flexibly, extracting the informative noise features from input faces and fusing them with high-level semantic features. Besides, noise-based fingerprints can emphasize high-frequency clues and expose more spatial local artifacts, which are largely ignored in the high-level semantic feature. Therefore, the noise feature plays a complementary role to the high-level semantic feature, and the two-stream learning strategy can be conducted by mining high-level and low-level artifacts in the feature space. 

As defined in Eq.~(1), the objective function of overall framework consists of the following three components: the classification loss $L_{c}$, noise prediction loss $L_{n}$, and binary mask prediction loss $L_{b}$, and ${\lambda}_{1}$ and ${\lambda}_{2}$ are hyper-parameters to weigh the loss components. To better demonstrate the effectiveness of the noise prediction module, we conduct an ablation study for the effects of loss weights ${\lambda}_{1}$ and ${\lambda}_{2}$. We report the detection results in Table.~\ref{loss_weights}. We can see that when the overall framework is trained without noise loss component ($i.e.$, $\lambda_{1}$ = 0), no matter how we fine-tune the value of $\lambda_{2}$ from 0.1 to 10.0, the manipulation detection performances are consistently lower than the model trained with the noise loss component $L_{n}$. This experiment indicates that the adoption of noise information does improve the framework's face manipulation detection performance.

To further demonstrate the effectiveness of devised noise map prediction module, we visualize the predicted noise maps, input faces, and ground-truth noise maps in Fig.~\ref{Noise_pred}. The top row shows the input faces. The middle and bottom rows present the corresponding ground-truth and predicted noise maps. It can be observed that the noise predictor can efficiently predict the noise maps, demonstrating that our model has learned the low-level noise information. We further use red boxes to mark the local artifacts in the predicted noise maps, and these clues are hard to be recognized in the RGB space. We can conclude that, in this model, the noise features carrying informative low-level local clues have been successfully extracted, thus leading to a certain face manipulation detection performance gain.

\subsubsection{Effects of different noise filters} 
To further explore the impacts of other noise filters, we conduct more experiments using the SRM high-pass filter and the CNN-based filter. Table~\ref{eval_filter} also reports the detection results of directly taking the noise map as input for reference. SRM has been demonstrated its effectiveness in image forensics tasks \cite{wu2019mantra, zhou2018learning} as it can expose the high-frequency artifacts in the manipulated images. Moreover, the recently proposed CNN-based filter \cite{tian2020attention} has been widely used in various image denoising tasks. Compared with results of directing use noise as input, our framework exploiting the noise fingerprints in the feature space has a better detection performance, as shown in the last row in Table~\ref{eval_filter}. Furthermore, the results of using the three different noise filters achieve competitive forgery detection performances. 

\begin{table}
  \caption{Ablation study for the effects of loss weights ${\lambda}_{1}$ and ${\lambda}_{2}$.}
  \label{loss_weights}
  \centering
  \renewcommand\arraystretch{1.15}
  \scalebox{1.0}{\begin{tabular}{|c|c|c|c|c|c|}
    \hline
    {$\lambda_2$}  & {$\lambda_1$} & AUC(\%) &  AP(\%) & EER(\%)  \\
    \hline
    0.0 & - & 84.38 & 94.63 & 22.97\\
    \hline
    0.1 & - & 84.91 & 95.33 & 23.34\\
    \hline
    0.5 & - & 86.35 & 95.63 & 21.72 \\
    \hline
    1.0 & - & 86.87 & 95.98 & 21.03 \\
    \hline
    5.0 & - & 86.76 & 96.09 & 21.32\\
    \hline
    10.0 & - & 86.72 & 95.90 & 21.24 \\
    \hline
    1.0 & $\surd$ & \bm{$87.10$} & \bm{$96.18$} & \bm{$20.94$} \\
    \hline
\end{tabular}}
\end{table}

\begin{table}
  \caption{Detection performance using different noise filters.}
  \label{eval_filter}
  \centering
  \renewcommand\arraystretch{1.15}
  \scalebox{0.75}{\begin{tabular}{|c|c|c|c|c|c|c|}
    \hline
    Method  & AUC(\%) $\uparrow$ & ACC(\%) $\uparrow$ & AP(\%) $\uparrow$ & FPR(\%) $\downarrow$ & FNR(\%) $\downarrow$ & EER(\%) $\downarrow$ \\
    \hline
    Noise input  & 82.07 & 81.14 & 94.42 & 47.79 & 11.63 & 25.71 \\
    \hline
    SRM HP \cite{fridrich2012rich} & 81.38 & 81.95 & 93.67 & 51.61 & 9.66 & 26.19 \\
    \hline
    CNN-based \cite{tian2020attention} & 82.66 & 82.63 & 94.40 & 49.74 & 9.27 & 24.84 \\
    \hline
    Ours & 82.15 & 82.75 & 94.55 & 53.63 & 8.15 & 25.66 \\
    \hline
\end{tabular}}
\end{table}

\subsubsection{Discussion}
Herein, we conduct a series of ablation studies to evaluate the effectiveness of the dedicated learning strategy and framework architecture. We also study the impacts of feature aggregation strategy, different backbones, noise features, and different noise filters to comprehensively interpret the proposed framework. The qualitative and quantitative experimental results demonstrate that the designed sub-modules and the multi-scale learning strategy lead the model to mine more artifact clues from the input faces, which further improves the final forgery detection performance.  

\section{Conclusions and Future Work}

This paper presented a novel framework to tackle the face manipulation problem. In particular, we introduced two complementary tasks, including semantic map prediction and noise map prediction, to capture the semantic-level and noise-level information to perform both face manipulation detection and forgery localization.  Extensive experimental results show that the proposed two-stream multi-scale framework outperforms the state-of-the-art detection methods and state-of-the-art cross-dataset detection methods for both high-quality and low-quality faces. 

The proposed semantic map prediction module enables the model to perform face manipulation localization, and it also constrains the model to focus on manipulated regions, thus leading to a better binary and multi-class classification performance. Furthermore, the noise map prediction module serves as a complementary module, and it provides significative noise-level clues and subsequently empowers the final decision-making. 

Last but not least, an ablation study demonstrated the effectiveness of the semantic map prediction module and the noise map prediction module, and the multi-scale feature learning strategy indeed helps the model improve its manipulation detection performance. 

While our proposed method has shown to be effective for face manipulation detection and localization tasks, we limit our scope to only four FF++ attacks (e.g., DeepFakes, Face2Face, FaceSwap, NeuralTextures) and two GAN attacks (attribute-manipulated faces and entirely synthetic faces).  Therefore, adapting our method to unseen face manipulation is worth investigating in the future, although we already provided a good indication of the method's performance considering the complex cross-dataset validation. On the other hand, since natural  videos also face a similar threat of malicious manipulation, it is also worth applying our method to the manipulation detection based on other types of multimedia content in the future. 

\section{Acknowledgement}

This work is supported in part by Shenzhen Virtural University Park, The Science Technology and Innovation Committee of Shenzhen Municipality (Project No: 2021Szvup128).

A. Rocha thanks the financial support of the S\~ao Paulo Research Foundation for grant D\'ej\`aVu \#2017/12646-3.

This work is also supported in part by the National Natural Science Foundation of China under 62022002, in part by the Hong Kong Research Grants Council General Research Fund (GRF) under Grant 11203220.
\ifCLASSOPTIONcaptionsoff
  \newpage
\fi

\bibliographystyle{IEEEtran}
\bibliography{main}

\end{document}